\documentclass[10pt,twocolumn,letterpaper]{article}

\usepackage[pagenumbers]{cvpr} 

\usepackage[accsupp]{axessibility}
\usepackage{graphicx}
\usepackage{amsmath}
\usepackage{amssymb}
\usepackage{booktabs}

\usepackage{multirow}
\usepackage{bigstrut}
\usepackage{array}

\usepackage[pagebackref,breaklinks,colorlinks]{hyperref}

\usepackage[capitalize]{cleveref}
\crefname{section}{Sec.}{Secs.}
\Crefname{section}{Section}{Sections}
\Crefname{table}{Table}{Tables}
\crefname{table}{Tab.}{Tabs.}

\def\cvprPaperID{7769} 
\def\confName{CVPR}
\def\confYear{2023}

\begin{document}

\title{Efficient Map Sparsification Based on 2D and 3D Discretized Grids}

\author{
  Xiaoyu Zhang \qquad Yun-Hui Liu\\
  T Stone Robotics Institute, Chinese University of Hong Kong \\
  Hong Kong Centre for Logistics Robotics\\
  {\tt\small zhang.xy@link.cuhk.edu.hk \qquad yhliu@mae.cuhk.edu.hk}
}
\maketitle

\begin{abstract}
  Localization in a pre-built map is a basic technique for robot autonomous navigation. Existing mapping and localization methods commonly work well in small-scale environments. As a map grows larger, however, more memory is required and localization becomes inefficient. To solve these problems, map sparsification becomes a practical necessity to acquire a subset of the original map for localization. Previous map sparsification methods add a quadratic term in mixed-integer programming to enforce a uniform distribution of selected landmarks, which requires high memory capacity and heavy computation. In this paper, we formulate map sparsification in an efficient linear form and select uniformly distributed landmarks based on 2D discretized grids. Furthermore, to reduce the influence of different spatial distributions between the mapping and query sequences, which is not considered in previous methods, we also introduce a space constraint term based on 3D discretized grids. The exhaustive experiments in different datasets demonstrate the superiority of the proposed methods in both efficiency and localization performance. The relevant codes will be released at \url{https://github.com/fishmarch/SLAM_Map\_Compression}.
\end{abstract}

\section{Introduction}
\label{sec:intro}
To realize autonomous navigation for robots, localization in a pre-built map is a basic technique. Lots of algorithms have been proposed for mapping and localization using different sensors, including camera \cite{Mur-Artal2017} and lidar \cite{zhang2014loam}. These algorithms commonly work well in some small-scale environments now. When applied in large-scale environments or in long-term, however, new challenges appear and need to be settled for practical applications. Some of these problems are time and memory consuming.

When using cameras, visual simultaneous localization and mapping (SLAM) is a commonly used method to build maps. In visual SLAM, redundant features are extracted from images and then constructed as landmarks in a map, such that camera poses can be tracked robustly and accurately. These redundant landmarks promise good localization results in small-scale environments. As more and more images are received when working in large-scale environments or in long-term, however, memory consumption is increasing unboundedly. Localizing in such large maps will also be more time-consuming. These problems are especially severe for some low-cost robots.

\begin{figure}[t]
	\centering
  \includegraphics[width=7.5cm]{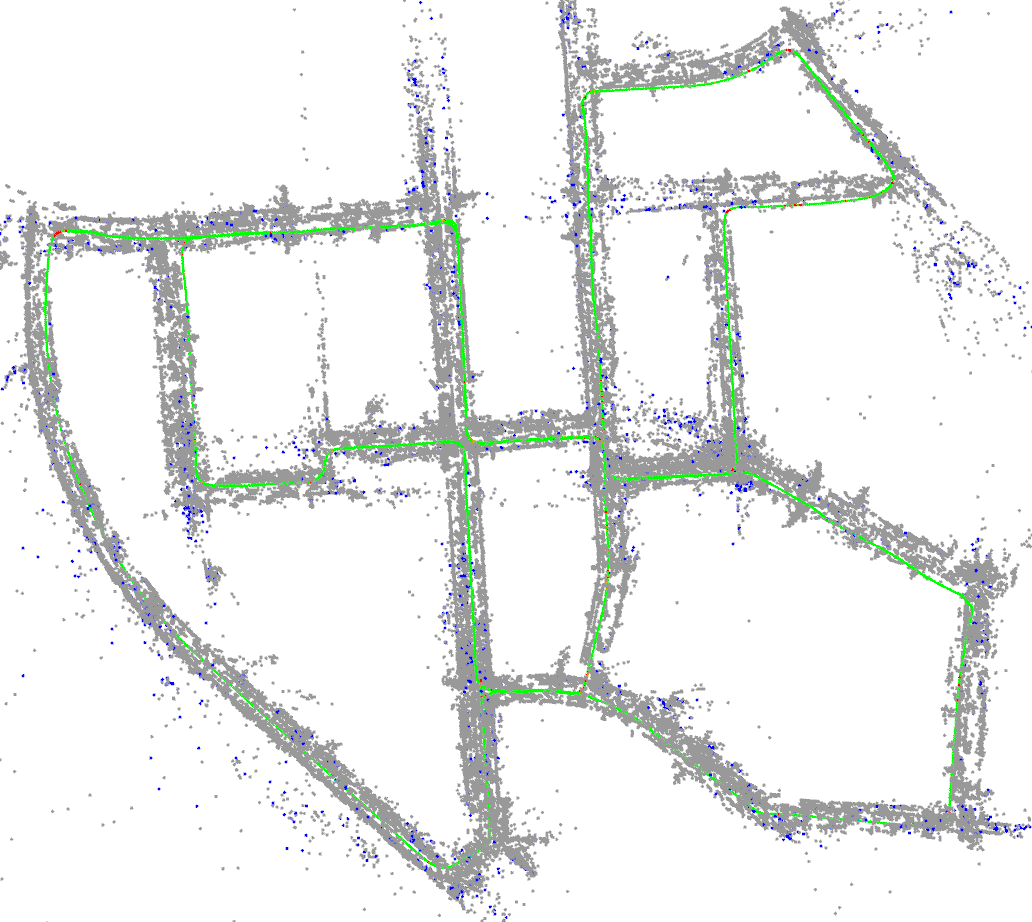}
  \captionsetup{font=small}
	\caption{Localization results in a compact map. The original map is constructed from sequence 00 of the KITTI dataset \cite{geiger2012we}. The original map consists of $141$K landmarks, indicated by gray points; while the compact map consists of only $5$K landmarks, indicated by blue points. $96.15\%$ of the query images are localized successfully in the compact map, indicted by green poses.}
  \vspace{-1em}
	\label{kitti}
\end{figure}

Actually, not all landmarks are necessary for robots to localize in pre-built maps. In theory, even only 4 matched landmarks can determine a camera pose using EPnP \cite{lepetit2009epnp} (more landmarks are commonly used for robust and accurate estimation). This reveals that maps can be compressed and still retain comparable performance for localization. Map compression can be classified into two types: descriptor compression \cite{lynen2015get, Sattler_2015_ICCV, Cheng_2019_ICCV} and landmark sparsification \cite{li2010location, Park_2013_CVPR_Workshops}. The research of this paper falls in the latter one, which is to find a subset of an original map while maintaining comparable localization performance. A subset map is called compact map in this paper. For example, as indicated in ~\cref{kitti}, only $3.91\%$ of the original landmarks are selected as the compact map, in which more than $96\%$ of the query images can still be localized successfully.

To select an optimal subset for localization, map sparsification is related to a $K$-cover problem \cite{li2010location}, which means the number of landmarks in a compact map is minimized while keeping the number of associated landmarks in each image larger than a threshold (i.e., $K$). To solve the $K$-cover problem, it can be formulated as mixed-integer linear programming, through which an optimal subset is obtained. The original formulation only considers the number of landmarks for localization, while their distribution also affects localization performance. Therefore, some works design and add quadratic terms, formulating mixed-integer quadratic programming to enforce a more uniform distribution of selected landmarks \cite{Park_2013_CVPR_Workshops, dymczyk2015keep}. However, these quadratic terms slow down the optimization speed heavily. The required high memory capacity and heavy computation become severe limitations of these map sparsification methods. For example, in our experiments, the mixed-integer quadratic programming methods cannot be used for the maps containing more than 55K landmarks, because all computer memory has been consumed.

To select uniformly distributed landmarks and at the same time maintain the computation efficiency, we keep map sparsification formulation in a linear form in this paper. We firstly discretize images into 2D fix-sized grids. Then for all observed landmarks, we can find which cells they fall in. Therefore, more occupied cells reflect a more uniform distribution of landmarks. This can be formulated in a linear form easily. In this way, uniformly distributed landmarks are selected efficiently, and thus the localization performance in compact maps will be better. 

Another severe limitation is that all of past works assume the spatial distribution of the query sequence is close to that of the mapping sequence. Then landmarks are selected only based on their association with the images of the mapping sequence. However, this assumption cannot be guaranteed in real robotic applications. The perspective difference between query and mapping sequences may cause the localization in compact maps to fail unexpectedly.

To ensure more query images from whole 3D space can be localized successfully in compact maps, we propose to select landmarks based on not only their association with mapping images but also their visibility in 3D space. The visibility region of a landmark is defined based on its viewing angle and distance. The 3D space is also discretized into a 3D discretized grid. For each 3D cell, all visible landmarks are collected, and a constraint on the number of visible landmarks is added into map sparsification formulation. In this way, landmarks are selected to maintain localization performance for query images from the whole space. 

In summary, the contributions of this paper are as follows: 1) We propose an efficient map sparsification method formulating uniform landmark distribution in a linear form to keep the computation efficiency. 2) We propose to perform map sparsification involving the visibility of landmarks to achieve better localization results for the query images from the whole space. 3) We conduct exhaustive experiments in different datasets and compare with other state-of-the-art methods, showing the effectiveness and superiority of our methods for different kinds of query sequences.

\section{Related Work}
Map compression or sparsification has received many research attention in SLAM and structure from motion (SfM) because of the burden of maintaining and utilizing a large map. Different methods are also proposed to solve this problem. In \cite{lynen2015get, Sattler_2015_ICCV, Cheng_2019_ICCV}, the descriptors are quantized to compact forms to reduce memory consumption. Mera-Trujillo et al. \cite{9320448} propose a constrained quadratic program method and design an efficient solver to select landmarks. Contreras et al. \cite{7353365} develop a map compression method based on traveled trajectories, and they also propose an online system performing SLAM and map compression simultaneously \cite{7989523}. Some SLAM systems \cite{Zhang_2015_CVPR, 8964284, 7139839} are also designed to select and build a compact map online, but much more landmarks are still remained compared with the offline methods. Recently, some learning-based map sparsification methods have also appeared and achieved good results \cite{Yang_2022_CVPR, Chang22cvpr}, but they require high-cost GPUs to train and struggle to be applied in different unseen scenes.  

Map sparsification is more commonly regarded as a $K$-cover problem. Since $K$-cover is an NP-hard problem, Li et al. \cite{li2010location} propose to use a greedy algorithm to select the landmark observed by the largest number of not-yet-fully covered frames.  Cao et al. \cite{cao2014minimal} also use a greedy algorithm to select landmarks, considering both the coverage and distinctiveness. In \cite{7725940}, the original $K$-cover method is improved by predicting the parameter $K$. Similarly, the landmarks are scored and sampled based on the observation count, covariance, and descriptor stability in \cite{dymczyk2015gist}. Moreover, the authors come up with more scoring factors to rank landmarks for selecting in \cite{7759673}. These methods can provide compact maps for localization, but depend on a careful design of selecting methods and cannot get the optimal solution.

Another $K$-cover based map sparsification solution is to be formulated as a mixed-integer programming problem, which is introduced in detail in \cite{Park_2013_CVPR_Workshops}. The linear formulation does not consider the distribution of the selected landmarks but can be solved efficiently, and therefore it is adopted in some later works \cite{7759673, 8354218}. The mixed-integer quadratic programming formulation encourages the uniformly distributed landmarks but requires a high capacity of memory and heavy computation \cite{Park_2013_CVPR_Workshops} and becomes impractical for large maps. To improve the computation efficiency, Dymczyk et al. \cite{dymczyk2015keep} have to divide the whole map into several small parts and perform map sparsification separately. Camposeco et al. \cite{Camposeco_2019_CVPR} keep the linear form of map sparsification but divide each image into $q$ uniformly-sized cells and let each cell be covered by $K/q$ landmarks. This method seems reasonable, but we find in the experiments that the number of corresponding selected landmarks in some images is much less than $K$ actually, and thus the localization results get worse. Similarly, the proposed method also makes use of discretized images, but the image is still considered as a whole and thus avoids the above problems. 

All of the above methods are based on the existing association between landmarks and images, thus landmarks are selected only to ensure localization performance of these images. For an arbitrary query image with a different viewing perspective, localization may fail unexpectedly, especially when the compression ratio is high. To solve this problem, the proposed method considers landmark visibility in the 3D space and adds constraints to preserve enough visible landmarks for the whole space in compact maps.

\section{Map Sparsification Formulation}
\label{sec:formulation}
The proposed method is designed mainly for the maps built from feature-based visual SLAM systems (e.g., ORB-SLAM \cite{Mur-Artal2017}). Receiving sequences of images, the feature points observed by keyframes are constructed as landmarks. Thus, an original map $\mathcal{M}$ consists of landmarks $\left\{\boldsymbol{p}_i\right\}_{i = 1,2,\dots,N}$, keyframes $\left\{\boldsymbol{F}_j\right\}_{j = 1,2,\dots,M}$, and their association. The map sparsification aims to decrease the number of landmarks. We firstly introduce commonly used formulations of map sparsification in past works \cite{Park_2013_CVPR_Workshops}.

Using mixed-integer programming, map sparsification can be formulated as:
\begin{equation}
  \label{eq:slack}
  \begin{array}{cl}
  \underset{\mathbf{x}, \boldsymbol{\xi}}{\operatorname{minimize}} & \mathbf{q}^{\top} \mathbf{x}+\lambda \mathbf{1}^{\top} \boldsymbol{\xi} \\
  & \mathbf{Ax} \geq K \mathbf{1}-\boldsymbol{\xi} \\
  \operatorname {subject \  to}
  & \mathbf{x} \in \left\{0, 1\right\}^{N} \\
  & \boldsymbol{\xi} \in\left\{0, \mathbb{Z}_{+}\right\}^{M}
  \end{array}
\end{equation}
where $\mathbf{x}$ is a binary vector whose $i^{\operatorname{th}}$ element indicates whether the landmark $\boldsymbol{p}_i$ is selected or not. $\mathbf{q}$ is a weight vector to select landmarks. $\mathbf{A}$ is a binary matrix describing the association between landmarks and keyframes, i.e., the element $\mathbf{A}_{ij}$ indicates whether the landmark $\boldsymbol{p}_j$ is visible in the keyframe $\boldsymbol{F}_i$. $K$ is the desired minimum number of visible landmarks in each keyframe and can be used to adjust compression ratio. To handle the situation when the total number of associated landmarks of a keyframe is less than $K$, a slack variable $\boldsymbol{\xi}$ is used to relax the hard constraints, and $\lambda$ determines the hardness of the constraints.

Therefore, \cref{eq:slack} describes the mixed-integer linear programming method for map sparsification. The distribution of selected landmarks is not considered in this method. To involve the distribution of landmarks, quadratic terms are commonly used:
\begin{equation}
  \label{eq:quadratic}
  \begin{array}{cl}
  \underset{\mathbf{x}, \boldsymbol{\xi}}{\operatorname{minimize}} & \frac{1}{2} \mathbf{x}^{\top} \mathbf{Q} \mathbf{x}+\mathbf{q}^{\top} \mathbf{x}+\lambda \mathbf{1}^{\top} \boldsymbol{\xi} \\
  & \mathbf{Ax} \geq K \mathbf{1}-\boldsymbol{\xi} \\
  \operatorname {subject \  to}
  & \mathbf{x} \in \left\{0, 1\right\}^{N} \\
  & \boldsymbol{\xi} \in\left\{0, \mathbb{Z}_{+}\right\}^{M}
  \end{array}
\end{equation}
where $\mathbf{Q}$ is a symmetric matrix. The element $\mathbf{Q}_{ij}$ indicates the relation between $\boldsymbol{p}_i$ and $\boldsymbol{p}_j$. For example, it can be the number of common observations for each pair of landmarks. Therefore, the landmarks that are observed together with others frequently would be discarded. As a result, selected landmarks are distributed more uniformly. \cref{eq:quadratic} describes the commonly used map sparsification formulation in past works, which considers both the number and distribution of landmarks for localization.

\section{Map Sparsification Based on 2D Grids}
Uniformly distributed landmarks commonly provide better localization results. Because of the quadratic terms, however, the computation complexity increases largely. Besides, \cref{eq:quadratic} encourages selection of rarely observed landmarks, which may have larger errors and be detrimental for localization \cite{dymczyk2015keep}. 

\begin{figure}[t]
	\centering
    \includegraphics[width=7cm]{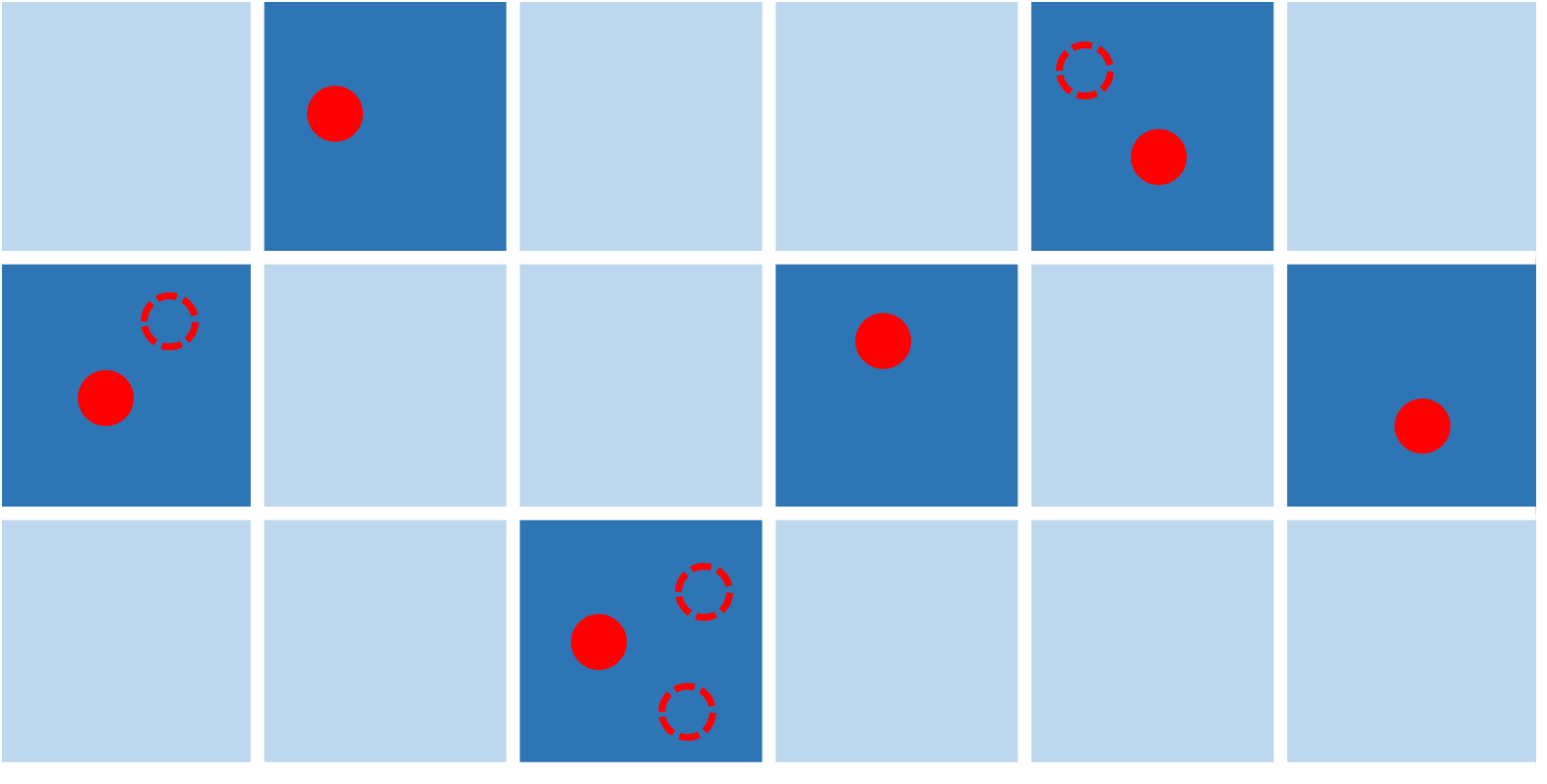}
    \captionsetup{font=small}
	\caption{Illustration of discretized images and associated landmarks. The \emph{occupied} cells are drawn in deeper color. The red points represent the projected position of associated landmarks. Dotted points can be discarded without changing occupancy.}
  \vspace{-1em}
	\label{cells}
\end{figure}

To solve these problems, we design a linear term to encourage a uniform distribution of selected landmarks. As illustrated in Fig.~\ref{cells}, the image is discretized into fixed-size grids (e.g., $C\times{R}$ cells). Each cell has two stats: \emph{empty} or \emph{occupied}. Based on the camera model, projected positions of landmarks can be computed easily. Then a cell is \emph{occupied} if at least one landmark is projected into it; otherwise, it is \emph{empty}. Therefore, to acquire uniformly distributed landmarks, we encourage selecting the landmarks that make more cells to be \emph{occupied}. This concept can be easily formulated into \cref{eq:slack}, which becomes:
\begin{equation}
  \begin{array}{cl}
  \underset{\mathbf{x}, \boldsymbol{\xi}, \boldsymbol{\phi}}{\operatorname{minimize}} & \mathbf{q}^{\top}\mathbf{x} +\lambda_1\mathbf{1}^{\top}\boldsymbol{\xi} + \lambda_2\mathbf{1}^{\top}\boldsymbol{\phi}\\
  & \mathbf{Ax} \geq K \mathbf{1} -\boldsymbol{\xi} \\
  & \mathbf{Bx} \geq \mathbf{1} - \boldsymbol{\phi} \\
  \operatorname {subject \  to}
  & \mathbf{x} \in \left\{0, 1\right\}^{N} \\
  & \boldsymbol{\xi} \in\left\{0, \mathbb{Z}_{+}\right\}^{M} \\
  & \boldsymbol{\phi} \in \left\{0, 1\right\}^{M\times C \times R} \\
  \end{array}
  \label{eq:ours1}
\end{equation}
where $\mathbf{B}$ is a binary matrix describing the association between the landmarks and all cells, i.e., the element $\mathbf{B}_{ij}$ indicates whether the projected position of the landmark $\boldsymbol{p}_j$ falls within the cell $\boldsymbol{c}_i$. We encourage every cell to keep at least one corresponding landmark and use the slack variable $\boldsymbol{\phi}$ to relax this constraint. An example is illustrated in \cref{cells}, the dotted points can be discarded without changing the number of \emph{occupied} cells.

\cref{eq:ours1} formulates map sparsification as a mixed-integer linear programming problem and thus can be solved efficiently. The selected landmarks are directly encouraged to spread uniformly in all corresponding keyframes. Thus better localization performance will be achieved.

\section{Map Sparsification In 3D Space}
From the previous section, it is noticed that map sparsification is formulated based on the association between landmarks and keyframes. This association can be easily constructed from feature matching during mapping. All past works are based on such association and assume the spatial distribution of the query sequence will be close to that of the mapping sequence. Thus selected landmarks can be matched again in query images. Obviously, this assumption cannot be promised in real robotic applications. Some selected landmarks would not be matched in query images because of different viewing perspectives, and thus, the localization may fail unexpectedly. 

In this section, we will introduce our proposed map sparsification method that involves landmark visibility in 3D space and thus improves localization performance for query images from the whole space. 

\subsection{Landmark Visibility}
Feature points (e.g., ORB \cite{rublee2011orb}) are commonly extracted and constructed as landmarks in visual SLAM. One of the basic requirements of used features is robust matching from different viewing perspectives. In other words, it is assumed that a valid feature point is visible and can be matched within a certain region. Specifically, we assume a feature point can be matched when the viewing angle and distance are within certain thresholds. Such specific regions can be approximated from mapping process. Similarly defined visibility is also used for efficient feature matching \cite{Mur-Artal2017}. Deng et al. \cite{deng2018feature} also predict such feature matching to ensure localization performance in an active SLAM framework.

For a landmark $\boldsymbol{p}$ observed by multiple keyframes, an average viewing direction ${\Bar{\boldsymbol v}^{p}}$ can be computed from all viewing directions, pointing from $\boldsymbol{p}$ to the optical centers of these keyframes. Therefore, the first condition for $\boldsymbol{p}$ to be visible is that the viewing angle is within the threshold,
\begin{equation}
    \label{eq:cond1}
    \langle \boldsymbol{\Bar{v}}^{p}, \boldsymbol{v}^{i} \rangle < \theta_{th}^p
\end{equation}
where $\boldsymbol{v}^{i}$ is the viewing direction for a query position. In our experiments, $\theta_{th}^p$ is set according to the largest viewing angle during mapping.

Besides, an image pyramid is commonly employed for matching features from different distance. Therefore, the second condition for $\boldsymbol{p}$ to be visible is that the viewing distance $d_i^p$ from a query position is within the following range,
\begin{equation}
  \label{eq:cond2}
  d_{\min}^p < d_i^p < d_{\max}^p
\end{equation}
where $d_{\min}^p$ and $d_{\max}^p$ are also set according to the smallest and largest viewing distance during mapping.

\begin{figure}[t]
	\centering
  \includegraphics[width=5cm]{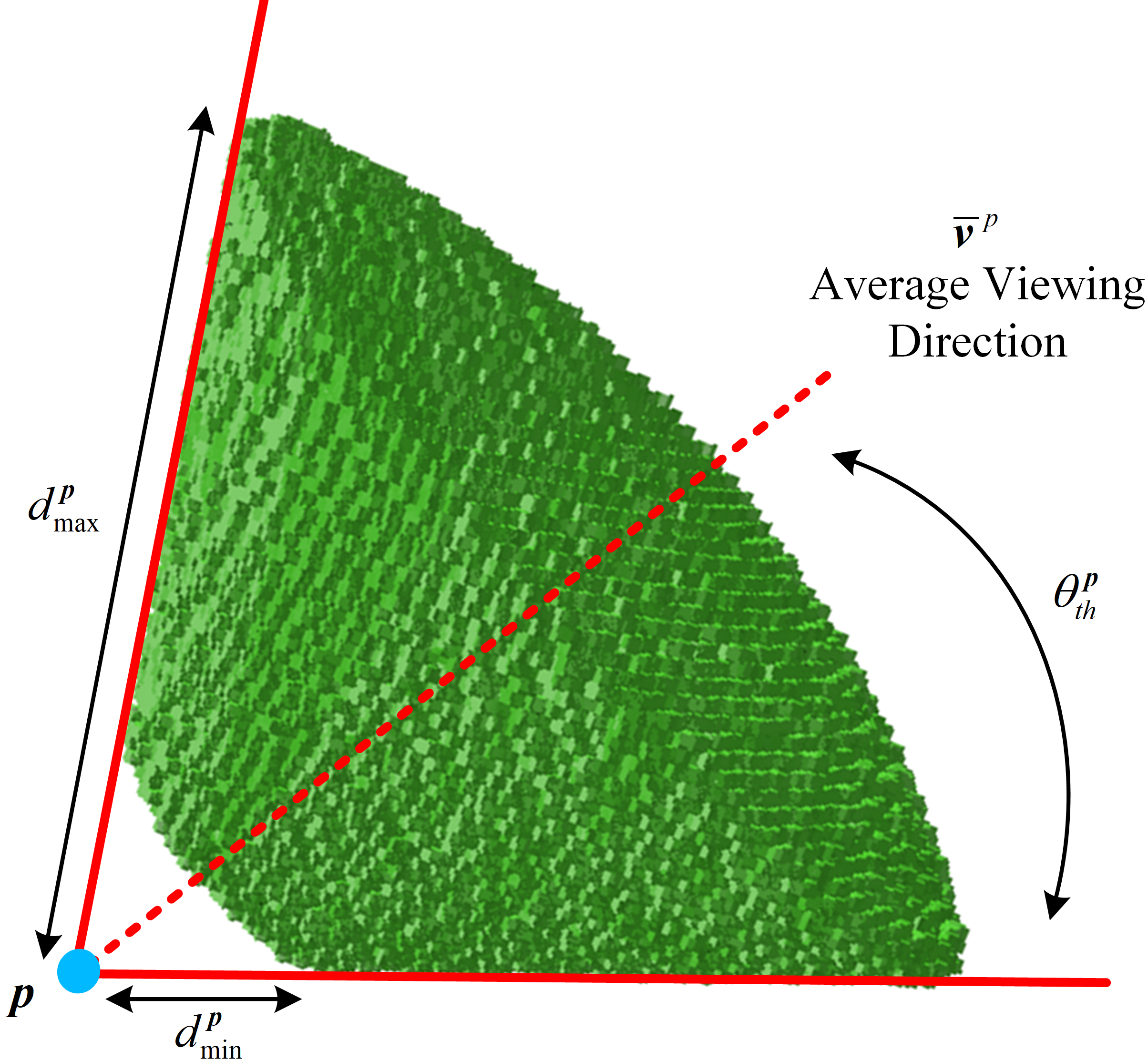}
  \captionsetup{font=small}
	\caption{3D cells in which the landmark $\boldsymbol{p}$ is visible.}
  \vspace{-1em}
	\label{scope}
\end{figure}

Therefore, a landmark is assumed to be visible when meeting the conditions \cref{eq:cond1} and \cref{eq:cond2}. The visible region forms a truncated spherical cone, as indicated in \cref{scope}.

\subsection{Map Sparsification Based on 3D Grid}
For map sparsification methods described in \cref{sec:formulation}, enough visible landmarks are preserved for every keyframe based on the constraint $\mathbf{Ax} \geq K \mathbf{1} -\boldsymbol{\xi}$. Therefore, the compact map can be used to localize the query images that are close to the mapping sequence as expected. However, for query images taken from different viewing perspectives, which is common in practical robotic applications, the number of matched visible landmarks may decrease, and thus the localization may fail unexpectedly. 

To solve the above problem, we propose a map sparsification method considering landmark visibility in 3D space. We firstly discretize the whole space into a 3D grid. The corresponding 3D cells from which it is visible are collected for each landmark, as indicated in \cref{scope}. Therefore, each cell stores all visible landmarks from that position. To decrease computation, we define a threshold $K_2$, and the 3D cell is \emph{valid} only if the number of corresponding visible landmarks is larger than $K_2$.

In the proposed map sparsification method, landmarks are selected to keep as many \emph{valid} 3D cells as possible. Therefore, an extra constraint is introduced into the formulation for each valid 3D cell: 
\begin{equation}
  \begin{array}{cl}
  \underset{\mathbf{x}, \boldsymbol{\xi}, \boldsymbol{\phi}, \boldsymbol{\varphi}}{\operatorname{minimize}} & \mathbf{q}^{\top}\mathbf{x} +\lambda_1\mathbf{1}^{\top}\boldsymbol{\xi} + \lambda_2\mathbf{1}^{\top}\boldsymbol{\phi} + \lambda_3\mathbf{1}^{\top}\boldsymbol{\varphi}\\
  & \mathbf{Ax} \geq K_1 \mathbf{1} -\boldsymbol{\xi} \\
  & \mathbf{Bx} \geq \mathbf{1} - \boldsymbol{\phi} \\
  & \mathbf{Cx} \geq K_2 \mathbf{1} -\boldsymbol{\varphi} \\
  \operatorname {subject \  to}
  & \mathbf{x} \in \left\{0, 1\right\}^{N} \\
  & \boldsymbol{\xi} \in\left\{0, \mathbb{Z}_{+}\right\}^{M} \\
  & \boldsymbol{\phi} \in \left\{0, 1\right\}^{M\times C \times R} \\
  & \boldsymbol{\varphi} \in\left\{0, \mathbb{Z}_{+}\right\}^{S}
  \end{array}
  \label{eq:ours2}
\end{equation}
where $\mathbf{C}$ is a binary matrix describing the association between landmarks and valid 3D cells, i.e., the element $\mathbf{C}_{ij}$ indicates whether the landmark $\boldsymbol{p}_j$ is visible for the 3D cell $\boldsymbol{g}_i$. As before, the slack variable $\boldsymbol{\varphi}$ is used to relax the constraints.

Therefore, the proposed map sparsification method aims to preserve enough visible landmarks not only for mapping keyframes but also for the whole 3D space. As a result, more query images from the whole space can be localized successfully.

\section{Experiments}
\label{sec:expr}
The proposed map sparsification based on 2D grids (i.e., \cref{eq:ours1}) is firstly compared with previous works to show its superiority in performance and efficiency for localizing query images close to mapping sequences. Then we demonstrate the effectiveness of map sparsification involving the visibility of landmarks in 3D space (i.e., \cref{eq:ours2}) and its superiority when query images are not close to mapping sequences. More experimental results can be found in the supplementary material, including visualization results and the comparison between original and compact maps in terms of memory consumption and localization efficiency.

\subsection{Experimental Setup}
All experiments are tested on a laptop with an i7-7700HQ 2.80 Hz CPU and 16GB RAM. The mixed-integer programming problem is solved using Gurobi\footnote{https://www.gurobi.com}.

The proposed methods are tested on three commonly used SLAM datasets: ICL-NUIM \cite{handa2014benchmark}, EuRoC \cite{burri2016euroc}, and KITTI \cite{geiger2012we}. These datasets cover indoor and outdoor, small-scale and large-scale environments; the number of landmarks in original maps changes from thousands to more than one hundred thousand. The original map is constructed by running a state-of-the-art visual SLAM system (i.e., ORB-SLAM2 \cite{Mur-Artal2017}). Query images are localized in the map based on the re-localization module in ORB-SLAM2, in which localization is successful when enough inliers are matched. Different map sparsification methods are compared in terms of localization rate and run-time. Localization rate is the ratio of the number of successfully localized images to the number of all query images in one sequence. 

The weight vector $\mathbf{q}$ in all methods is set based on the number of matched times for each landmark, just like in \cite{Park_2013_CVPR_Workshops}. The proposed method based only on 2D grids (i.e., \cref{eq:ours1}) is denoted as Ours-2D, while the method based on 2D and 3D grids (i.e., \cref{eq:ours2}) is denoted as Ours-3D. The proposed methods are compared with several state-of-the-art $K$-cover-based map sparsification methods. The map sparsification using mixed-integer linear programming (i.e., \cref{eq:slack}) is denoted as LP. For the methods using mixed-integer quadratic programming (i.e., \cref{eq:quadratic}), the weight matrix $\mathbf{Q}$ is set according to the number of co-observation times\cite{Park_2013_CVPR_Workshops} or average projected distance\cite{dymczyk2015keep}, and these two methods are denoted as QP1 and QP2 respectively. The map sparsification based on divided images \cite{Camposeco_2019_CVPR} is denoted as DI.

\subsection{Map Sparsification Based on 2D Grids}

\begin{table*}[t]
  \centering
  \caption{Comparison of different map sparsification methods in terms of localization rate. Num denotes the number of landmarks in the mam, Rate denotes the localization rate in the corresponding map. The highest localization rates in compact maps are labelled in bold. `-' means the method cannot work because of memory limit. }
    \begin{tabular}{p{0.1cm}p{0.7cm}<{\centering}|cp{1.0cm}<{\centering}|p{0.7cm}<{\centering}p{1.0cm}<{\centering}|p{0.7cm}<{\centering}p{1.0cm}<{\centering}|p{0.7cm}<{\centering}p{1.0cm}<{\centering}|p{0.7cm}<{\centering}p{1.0cm}<{\centering}|p{0.8cm}<{\centering}p{1.0cm}<{\centering}}
    \hline
    \multicolumn{2}{c|}{\multirow{2}[2]{*}{Dataset}} & \multicolumn{2}{c|}{Original Map} & \multicolumn{2}{c|}{Ours-2D} & \multicolumn{2}{c|}{LP \cite{Park_2013_CVPR_Workshops}} & \multicolumn{2}{c|}{QP1 \cite{Park_2013_CVPR_Workshops}} & \multicolumn{2}{c|}{QP2 \cite{dymczyk2015keep}} & \multicolumn{2}{c}{DI \cite{Camposeco_2019_CVPR}} \bigstrut[t]\\
    \multicolumn{2}{c|}{} &Num   & Rate  & Num   & Rate  & Num   & Rate  & Num   & Rate  & Num   & Rate  & Num   & Rate \bigstrut[b]\\
    \hline
          \multirow{8}[2]{*}{\rotatebox{90}{ICL-NUIM \cite{handa2014benchmark}}} & liv0  & 4061  & 98.34\% & 519   & \textbf{79.97\%} & 519   & {79.51\%} & 519   & 79.64\% & 519   & 78.65\% & 734   & 77.39\% \bigstrut[t]\\
          & liv1  & 4112  & 99.89\% & 445   & \textbf{87.47\%} & 445   & 86.23\% & 445   & 87.37\% & 445   & 85.51\% & 516   & 55.59\% \\
          & liv2  & 6435  & 100.0\% & 312   & \textbf{91.36\%} & 310   & 89.77\% & 310   & 90.57\% & 310   & 91.02\% & 833   & 89.89\% \\
          & liv3  & 6173  & 93.38\% & 532   & \textbf{76.21\%} & 531   & 73.71\% & 531   & 75.32\% & 531   & 73.55\% & 987   & 61.85\% \\
          & off0  & 5875  & 99.93\% & 398   & 95.62\% & 398   & 95.42\% & 398   & \textbf{95.76\%} & 398   & \textbf{95.76\%} & 791   & 84.88\% \\
          & off1  & 5992  & 93.57\% & 485   & \textbf{80.00\%} & 485   & 77.93\% & 485   & 79.07\% & 484   & 76.89\% & 736   & 37.20\% \\
          & off2  & 6838  & 100.0\% & 299   & \textbf{92.27\%} & 298   & 91.02\% & 298   & 90.34\% & 298   & 91.36\% & 759   & 90.45\% \\
          & off3  & 5367  & 99.52\% & 396   & \textbf{83.47\%} & 395   & 83.06\% & 395   & 78.31\% & 396   & 83.15\% & 430   & 51.05\% \bigstrut[b]\\
          \hline
          \multirow{11}[2]{*}{\rotatebox{90}{EuRoC \cite{burri2016euroc}}} & MH1   & 12506 & 99.04\% & 2978  & \textbf{97.33\%} & 2976  & 97.03\% & 2976  & 96.76\% & 2977  & 96.54\% & 4210  & 94.56\% \bigstrut[t]\\
          & MH2   & 11306 & 99.73\% & 2546  & 98.17\% & 2544  & 96.43\% & 2547  & \textbf{98.33\%} & 2547  & 98.30\% & 3665  & 93.30\% \\
          & MH3   & 15234 & 97.34\% & 2105  & 95.32\% & 2104  & 95.32\% & 2107  & \textbf{95.44\%} & 2105  & 95.17\% & 3720  & 91.30\% \\
          & MH4   & 17490 & 99.69\% & 744   & \textbf{97.37\%} & 743   & 97.32\% & 742   & {97.27\%} & 742   & 97.01\% & 2493  & 96.05\% \\
          & MH5   & 16956 & 99.77\% & 1116  & \textbf{98.87\%} & 1116  & 98.69\% & 1116  & 98.60\% & 1116  & 98.60\% & 2933  & 96.08\% \\
          & V101  & 6864  & 99.03\% & 461   & 86.04\% & 460   & 84.92\% & 461   & 85.52\% & 461   & \textbf{86.35\%} & 1011  & 79.14\% \\
          & V102  & 9878  & 98.12\% & 507   & \textbf{85.93\%} & 507   & 84.86\% & 506   & 85.87\% & 506   & 85.49\% & 1255  & 84.17\% \\
          & V103  & 11518 & 93.22\% & 1140  & 83.76\% & 1140  & \textbf{83.91\%} & 1140  & 83.09\% & 1140  & 83.05\% & 1661  & 77.89\% \\
          & V201  & 7346  & 94.69\% & 565   & \textbf{79.64\%} & 564   & 78.75\% & 564   & 78.62\% & 564   & 78.97\% & 990   & 78.75\% \\
          & V202  & 11791 & 99.48\% & 1254  & 93.25\% & 1254  & 92.86\% & 1254  & 93.07\% & 1254  & \textbf{93.42\%} & 2285  & 90.39\% \\
          & V203  & 14557 & 87.51\% & 1796  & \textbf{81.48\%} & 1795  & 81.27\% & 1796  & 81.11\% & 1795  & \textbf{81.48\%} & 2384  & 75.45\% \bigstrut[b]\\
          \hline
          \multirow{11}[2]{*}{\rotatebox{90}{KITTI \cite{geiger2012we}}} & 00    & 141673 & 100.0\% & 5536  & \textbf{96.15\%} & 5524  & 95.68\% & -     & -     & -     & -     & 18542 & 91.15\% \bigstrut[t]\\
          & 01    & 89443 & 91.01\% & 4218  & \textbf{76.02\%} & 4216  & \textbf{76.02\%} & -     & -     & -     & -     & 10204 & 71.84\% \\
          & 02    & 182499 & 99.96\% & 7660  & \textbf{97.47\%} & 7661  & 97.15\% & -     & -     & -     & -     & 25133 & 86.76\% \\
          & 03    & 24284 & 100.0\% & 837   & 96.00\% & 836   & 95.63\% & 837   & \textbf{96.13\%} & 836   & 95.76\% & 2952  & 88.76\% \\
          & 04    & 16824 & 100.0\% & 583   & \textbf{99.63\%} & 585   & 99.26\% & 572   & \textbf{99.63\%} & 572   & \textbf{99.63\%} & 2019  & 97.41\% \\
          & 05    & 74158 & 100.0\% & 2737  & \textbf{97.46\%} & 2738  & 97.25\% & -     & -     & -     & -     & 9655  & 90.62\% \\
          & 06    & 42522 & 100.0\% & 1879  & \textbf{98.27\%} & 1878  & 98.09\% & 1878  & 98.18\% & 1877  & 98.18\% & 5635  & 95.82\% \\
          & 07    & 27036 & 100.0\% & 1242  & 92.28\% & 1240  & 91.28\% & 1240  & 91.73\% & 1240  & \textbf{92.55\%} & 3517  & 92.01\% \\
          & 08    & 119323 & 100.0\% & 5903  & \textbf{94.87\%} & 5899  & 94.82\% & -     & -     & -     & -     & 16872 & 83.03\% \\
          & 09    & 55984 & 99.94\% & 2945  & 95.79\% & 2944  & 95.60\% & 2942  & \textbf{95.85\%} & 2943  & 95.79\% & 8224  & 88.37\% \\
          & 10    & 30933 & 99.92\% & 1729  & \textbf{97.09\%} & 1728  & 96.67\% & 1728  & 96.83\% & 1729  & 96.42\% & 4743  & 86.34\% \bigstrut[b]\\
        \hline
    \end{tabular}%
    \vspace{-1em}
    \label{tab_rate}%
\end{table*}%

Since previous methods assume the spatial distribution of the query sequence is close to that of the mapping sequence, one sequence is firstly used to build an original map and then the images from the same mapping sequence are re-localized in compact maps. In this subsection, the proposed map sparsification method is only based on 2D discretized grids (i.e., \cref{eq:ours1}). \cref{tab_rate} compares the localization rates in different maps obtained by different methods. An example of the selected sparse landmarks is illustrated by blue points in \cref{kitti}.

The experimental results demonstrate that the number of landmarks can be reduced largely while the localization is still successful for most query images. For example, only $4.37\%$ landmarks are selected from the map of sequence off2 and $3.46\%$ for sequence 04, but more than $90\%$ of the query images can still be localized successfully. The compression ratio is mainly controlled by the threshold $K$ in \cref{eq:slack}, \cref{eq:quadratic}, and \cref{eq:ours1}, and we use $K=50$ in our experiments. With a larger $K$, more landmarks will be selected in compact maps, and localization rate will be higher.  

For more sequences, our proposed method achieves the highest localization rate (labelled in bold), especially compared with the method only considering the number of landmarks for map sparsification (i.e., LP). Our method also gets slightly better results than other methods that consider landmark distribution (i.e., QP1, QP2, and DI). In QP1, the weight matrix $\mathbf{Q}$ is set according to the number of co-observation times between landmarks and as a result, rarely observed landmarks are preferred, which may be detrimental for localization \cite{dymczyk2015keep}. To overcome this limitation, QP2 sets the weight matrix $\mathbf{Q}$ based on the average distance between projected positions. But instead of uniform distribution, QP2 actually prefers to select landmarks that are projected far from the image center. Unlike these quadratic programming-based methods, our method formulates uniform distribution in a more clear and simple way, achieving better results.

DI gets unexpected bad performance in these testings. DI divides images into $q$ cells and forces each cell is covered by at least $K/q$ landmarks. Therefore, DI also achieves map sparsification in a linear programming form and considers both the number and distribution of landmarks. But we find in these testings that the slack variable $\lambda$ in \cref{eq:slack} needs to be small for DI to get a high compression ratio. As a result, the number of selected landmarks is much less than $K$ for some images, where localization would fail. Although based on similar discretized grids, our method still considers an image as a whole and thus avoids this problem.

\begin{table}[t]
  \centering
  \caption{Comparison of different map sparsification methods in terms of run-time and consumed memory. $N$ denotes the number of landmarks, $N_k$ denotes the number of keyframes.}
  \begin{tabular}{p{0.6cm}|cc|p{0.5cm}<{\centering}p{0.5cm}<{\centering}p{0.6cm}<{\centering}p{0.6cm}<{\centering}p{0.5cm}<{\centering}}
    \hline
    \multirow{2}[2]{*}{Seq.} & \multicolumn{2}{c|}{Original Map} & \multicolumn{5}{c}{Run-time (s)} \bigstrut[t]\\
          & $N$   & $N_k$   & Ours & LP    & QP1   & QP2   & DI \bigstrut[b]\\
    \hline
    liv0  & 4.1K  & 107   & 0.2 & 0.1 & 13.6 & 12.5 & 0.2 \bigstrut[t]\\
    off0  & 5.9K  & 133   & 0.3 & 0.1 & 27.9 & 18.9 & 0.2 \\
    V103  & 11.5K & 216   & 1.1 & 0.3  & 59.8 & 36.3 & 0.6 \\
    MH5   & 17.0K & 369   & 4.5 & 2.2 & 107 & 61.5 & 1.2 \\
    07    & 27.0K & 258   & 2.6  & 1.1  & 127 & 95.5 & 0.7 \\
    09    & 56.0K & 603   & 7.7  & 9.9  & 754 & 491 & 1.8 \\
    08    & 119K& 1269  & 33.2 & 19.2 & - & - & 4.9 \\
    02    & 182K& 1794  & 69.6 & 63.2 & - & - & 7.9\\
    \hline
    & $N$   & $N_k$   & \multicolumn{5}{c}{Consumed memory (MB)} \bigstrut[t]\\
    \hline
    liv0  & 4.1K  & 107   & 35 & 14 & 555 & 476 & 33 \bigstrut[t]\\
    off0  & 5.9K  & 133   & 61 & 27 & 842 & 708 & 21 \\
    V103  & 11.5K & 216   & 128 & 56  & 3048 & 1367 & 83 \\
    MH5   & 17.0K & 369   & 255 & 107 & 3532 & 1656 & 77 \\
    07    & 27.0K & 258   & 141  & 62  & 6315 & 5766 & 100 \\
    09    & 56.0K & 603   & 676  & 230  & 13005 & 10952 & 211 \\
    08    & 119K& 1269  & 1162 & 637 & - & - & 512 \\
    02    & 182K& 1794  & 3494 & 1502 & - & - & 1187 \bigstrut[b]\\
    \hline
    \end{tabular}%
  \label{tab_time}%
  \vspace{-1em}
\end{table}%

Another superiority of our method is computational efficiency. The comparison of run-time and memory consumption on different sequences is shown in ~\cref{tab_time}. The run-time for map sparsification depends on the number of landmarks, keyframes and associations among them. As the number of landmarks or keyframes increases, the run-time will commonly increase quickly. It is clear that the methods using quadratic programming are much slower, especially for the maps containing a large number of landmarks (e.g., sequence 09). In addition to solving more complex optimization problems, QP1 and QP2 also need much more time to set the weight matrix $\mathbf{Q}$, since the relation between every pair of landmarks needs to be found and this is not recorded in original maps. In our method, both the matrix $\mathbf{A}$ and $\mathbf{B}$ in \cref{eq:ours1} are set according to the relation between landmarks and keyframes, which can be retrieved from original maps directly and efficiently. In addition, since formulated in a linear form, our method can also be solved efficiently and is only slightly slower than LP. 

Similarly for memory consumption, ours consumes much less memory than quadratic programming. QP1 and QP2 cannot work for some sequences (e.g., 00) on our device with 16 GB memory. The testing for sequence 09 consumes nearly all of the memory when using QP1 and QP2, in which the original map contains 56K landmarks. 

Therefore, if the mapping sequence has covered most of the query space for localization, the proposed map sparsification based on 2D discretized grids can be used to decrease the number of landmarks to save memory and improve localization efficiency.

\subsection{Map Sparsification Based on 2D and 3D Grids}
\begin{figure}[t]
  \centering
  \includegraphics[width=8cm]{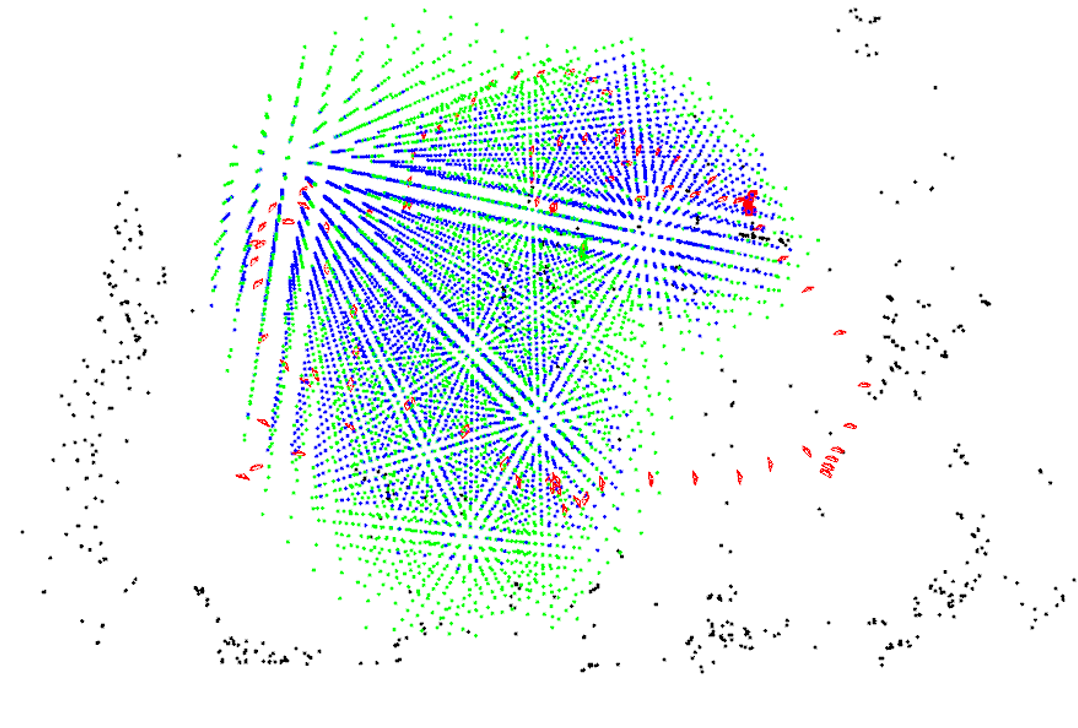}
  \captionsetup{font=small}
  \caption{Comparison of map sparsification results. The original map is constructed from sequence V201. There are 3401 blue points indicating valid 3D cells of the compact map acquired by Ours-2D; while there are 6468 green points indicating valid 3D cells of the compact map acquired by Ours-3D.}
  \vspace{-1em}
  \label{space}
\end{figure}

We have demonstrated that compact maps with much fewer landmarks achieve good localization performance when the spatial distribution of the query sequence is close to that of the mapping sequence. In this subsection, we continue to test the localization performance for query images from different distributions.

In ICL-NUIM \cite{handa2014benchmark} and EuRoC \cite{burri2016euroc} datasets, different sequences are recorded in the same environments. Therefore, one sequence is used to build original maps and then query images of \emph{other} sequences are also localized in compact maps. The comparison of experimental results is presented in Table~\ref{tab_diff}. For example, sequence off0 is used to build the map, and then sequence off0, off2 and off3 are used as query sequences to be localized in the original and compact maps. Not all sequences of these two datasets are used in this experiment since the successful localization is very low even in original maps for some sequences.


\begin{table*}[htbp]
  \centering
  \caption{Comparison of different map sparsification methods in terms of localization rate. Num denotes the number of landmarks in the map, Rate denotes the localization rate in the corresponding map. The sequence for mapping is indicated in bold and the images from several different sequences are localized in the map. The highest localization rates in compact maps are labelled in bold.}
    \begin{tabular}{c|c|ccc|c|cc|c|cc}
    \hline
     & \multicolumn{4}{c|}{\textbf{off0}}    & \multicolumn{3}{c|}{\textbf{off2}} & \multicolumn{3}{c}{\textbf{liv2}} \bigstrut\\
    \hline
    \multirow{2}[2]{*}{Map} & \multirow{2}[2]{*}{Num} & \multicolumn{3}{c|}{Rate} & \multirow{2}[2]{*}{Num} & \multicolumn{2}{c|}{Rate} & \multirow{2}[2]{*}{Num} & \multicolumn{2}{c}{Rate} \bigstrut[t]\\
          &       & off0  & off2  & off3  &              & off0  & off2  &              & liv1  & liv2 \bigstrut[b]\\
    \hline
    Original & 5875  & 99.93\% & 66.70\% & 52.26\% & 6838  & 65.65\% & 100.00\% & 6435  & 55.80\% & 100.00\% \bigstrut[t]\\
    Ours-2D & 398   & \textbf{95.62\%} & 27.95\% & 7.74\% & 299   & 4.31\% & \textbf{92.27\%} & 312   & 2.48\% & \textbf{91.36\%} \\
    Ours-3D & 399   & 95.49\% & \textbf{28.52\%} & \textbf{7.98\%} & 301 & \textbf{6.17\%} & 91.70\% & 314   & \textbf{4.04\%} & 90.68\% \\
    \hline
    & \multicolumn{4}{c|}{\textbf{MH1}}     & \multicolumn{3}{c|}{\textbf{MH4}} & \multicolumn{3}{c}{\textbf{MH5}} \bigstrut\\
    \hline
    \multirow{2}[2]{*}{Map} & \multirow{2}[2]{*}{Num} & \multicolumn{3}{c|}{Rate} & \multirow{2}[2]{*}{Num} & \multicolumn{2}{c|}{Rate} & \multirow{2}[2]{*}{Num} & \multicolumn{2}{c}{Rate} \bigstrut[t]\\
          &       & MH1   & MH2   & MH3   &              & MH4   & MH5   &             & MH4   & MH5 \bigstrut[b]\\
    \hline
    Original & 12506 & 99.04\% & 87.10\% & 52.53\% & 17490 & 99.69\% & 84.20\% & 16956 & 82.34\% & 99.77\% \bigstrut[t]\\
    Ours-2D & 2978  & \textbf{97.33\%} & 73.67\% & 45.76\% & 744   & \textbf{97.37\%} & 63.77\% & 1116  & 74.75\% & \textbf{98.87\%} \\
    Ours-3D & 2982  & 96.78\% & \textbf{73.97\%} & \textbf{46.18\%} & 745   & 97.06\% & \textbf{64.27\%} & 1121  & \textbf{75.71\%} & 98.74\% \\
    \hline
    & \multicolumn{4}{c|}{\textbf{MH2}}     & \multicolumn{3}{c|}{\textbf{V101}} & \multicolumn{3}{c}{\textbf{V102}} \bigstrut\\
    \hline
    \multirow{2}[2]{*}{Map} & \multirow{2}[2]{*}{Num} & \multicolumn{3}{c|}{Rate} & \multirow{2}[2]{*}{Num} & \multicolumn{2}{c|}{Rate} & \multirow{2}[2]{*}{Num} & \multicolumn{2}{c}{Rate} \bigstrut[t]\\
          &       & MH1   & MH2   & MH3   &          & V101  & V102  &              & V101  & V102 \bigstrut[b]\\
    \hline
    Original & 11306 & 91.78\% & 99.73\% & 51.81\% & 6864  & 99.03\% & 61.56\% & 9878  & 72.25\% & 98.12\% \bigstrut[t]\\
    Ours-2D & 2546  & 74.80\% & \textbf{98.17\%} & 35.50\% & 461   & \textbf{86.04\%} & 32.60\% & 507   & 34.61\% & 85.93\% \\
    Ours-3D & 2550  & \textbf{77.41\%} & 97.77\% & \textbf{38.69\%} & 462   & 85.79\% & \textbf{34.61\%} & 506   & \textbf{37.67\%} & \textbf{85.99\%} \\
    \hline
    & \multicolumn{4}{c|}{\textbf{MH3}}     & \multicolumn{3}{c|}{\textbf{V201}} & \multicolumn{3}{c}{\textbf{V202}} \bigstrut\\
    \hline
    \multirow{2}[1]{*}{Map} & \multirow{2}[1]{*}{Num} & \multicolumn{3}{c|}{Rate} & \multirow{2}[1]{*}{Num} & \multicolumn{2}{c|}{Rate} & \multirow{2}[1]{*}{Num} & \multicolumn{2}{c}{Rate} \bigstrut[t]\\
          &       & MH1   & MH2   & MH3   &              & V201  & V202  &              & V201  & V202 \\
    \hline
    Original & 15234 & 59.93\% & 52.80\% & 97.34\% & 7346  & 94.69\% & 52.77\% & 11791 & 75.22\% & 99.48\% \\
    Ours-2D & 2105  & 46.85\% & 37.23\% & \textbf{95.32\%} & 565   & 79.64\% & 19.48\% & 1254  & 49.82\% & 93.25\% \\
    Ours-3D & 2108  & \textbf{46.94\%} & \textbf{37.60\%} & 95.25\% & 569   & \textbf{80.00\%} & \textbf{22.77\%} & 1256  & \textbf{50.04\%} & \textbf{93.46\%} \\
    \hline
    \end{tabular}%
    \vspace{-1em}
  \label{tab_diff}%
\end{table*}%

As shown in \cref{tab_diff}, compact maps commonly achieve good localization performance for query images from mapping sequences, while localization rates may decrease largely for other query sequences. Because the map sparsification is mainly selecting landmarks to maintain the observations in keyframes of the mapping sequences, and thus discarding other redundant landmarks which may be useful for localizing query images from the whole space. Our proposed map sparsification method in \cref{eq:ours2} aims to add such extra constraints for the whole space. In most testings, ours-3D achieves higher localization rates for the query sequences that are different from mapping sequences. For the query images from the mapping sequence, ours-3D also achieves comparable localization performance with other map sparsification methods. Other map sparsification methods achieve similar results as Ours-2D, which are included in the supplementary material.

A comparison of compact maps acquired by ours-2D and ours-3D is illustrated in \cref{space}. The blue points indicate the valid 3D cells of ours-2D, while the green points indicate the valid 3D cells of ours-3D. The numbers of selected landmarks are similar using these two methods. But it is clear that much more valid 3D cells are kept using ours-3D and they are spreading over a larger space. As a result, more query images from the whole 3D space can be localized successfully. 
\begin{table}[t]
  \centering
  \caption{Comparison of average run-time (s) in different scenes.}
    \begin{tabular}{c|c|cc}
    \hline
    Scene     &  Resolution (m)     & Ours-2D & Ours-3D \bigstrut[t]\\
    \hline
    liv   & 0.15  & 0.25  & 1.00 \bigstrut[t]\\
    off   & 0.15  & 0.20  & 1.19 \\
    V1  & 0.20   & 0.99  & 8.06 \\
    V2  & 0.20   & 1.24  & 10.36 \\
    MH    & 0.40   & 3.62  & 11.34 \bigstrut[b]\\
    \hline
    \end{tabular}%
    \vspace{-1em}
  \label{tab_time_2}%
\end{table}%

Since Ours-3D is also formulated in a linear form, it can also be solved efficiently. \cref{tab_time_2} shows the average run-time in different scenes. The run-time of Ours-3D is related to the resolution of the 3D grids. In the experiment, we choose different resolutions for different scenes and thousands of valid cells are used for map sparsification. The run-time of Ours-3D does not increase much compared with Ours-2D, and Ours-3D is still much faster than QP1 and QP2. 

Therefore, if the mapping sequence has not covered the whole query space and query images may be taken with different viewing perspectives, the map sparsification involving the constraints from the 3D discretized grid is a better choice to select landmarks. 

\section{Conclusion}
In this paper, we propose two novel terms for efficient map sparsification. The first term is to enforce a uniform distribution of selected landmarks based on 2D discretized grids.  The second one adds a space constraint based on landmark visibility to weaken the assumption that the spatial distribution of the query sequence is close to that of the mapping sequence. Both two terms are formulated in efficient linear forms, and thus avoid heavy computation. The proposed terms can be chosen and set according to different conditions of query sequences. The exhaustive experiments have demonstrated the effectiveness and superiority of the proposed method, especially the computation efficiency compared with previous QP methods.

\vspace{0.5em}
\noindent
{{\bf Acknowledgments:} This work is supported by the CUHK T Sone Robotics Institute, and the InnoHK of the Government of Hong Kong via the Hong Kong Centre for Logistics Robotics.}

{\small
\bibliographystyle{ieee_fullname}
\bibliography{egbib}
}

\end{document}